\documentclass[11pt]{article}
\usepackage{acl}
\usepackage{times}
\usepackage{latexsym}
\usepackage[T1]{fontenc}
\usepackage[utf8]{inputenc}
\usepackage{microtype}
\usepackage{inconsolata}
\usepackage{graphicx}
\usepackage{booktabs}
\usepackage{multirow}
\usepackage{url}

\title{Towards High-Quality Machine Translation for Kokborok: A Low-Resource Tibeto-Burman Language of Northeast India}

\author{Badal Nyalang \\
  MWire Labs \\
  Shillong, Meghalaya, India \\
  \texttt{badal@mwirelabs.com} \\\And
  Biman Debbarma \\
  Department of Kokborok, \\
  Tripura University \\
  Agartala, Tripura, India \\
  \texttt{bimandblg@gmail.com} \\}

\begin{document}
\maketitle

\begin{abstract}
We present KokborokMT, a high-quality neural machine translation (NMT) system for Kokborok (ISO 639-3: \texttt{trp}), a Tibeto-Burman language spoken primarily in Tripura, India with approximately 1.5 million speakers. Despite its status as an official language of Tripura, Kokborok has remained severely under-resourced in the NLP community, with prior machine translation attempts limited to systems trained on small Bible-derived corpora achieving BLEU scores below 7. We fine-tune the NLLB-200-distilled-600M model on a multi-source parallel corpus comprising 36,052 sentence pairs: 9,284 professionally translated sentences from the SMOL dataset, 1,769 Bible-domain sentences from WMT shared task data, and 24,999 synthetic back-translated pairs generated via Gemini Flash from Tatoeba English source sentences. We introduce \texttt{trp\_Latn} as a new language token for Kokborok in the NLLB framework. Our best system achieves BLEU scores of 17.30 (en$\rightarrow$trp) and 38.56 (trp$\rightarrow$en) on held-out test sets, representing substantial improvements over prior published results. Human evaluation by three annotators yields mean adequacy of 3.74/5 and fluency of 3.70/5, with substantial agreement between trained evaluators ($\kappa = 0.67$). We will release the model, data, and code publicly under CC-BY-4.0 upon acceptance.
\end{abstract}

\section{Introduction}

Kokborok is one of the indigenous languages spoken by the Tiprasa people of Tripura, Northeastern India. The name itself is a compound of \textit{kok} (language) and \textit{borok} (people), literally meaning ``language of the people.'' With approximately 1.5 million speakers across Tripura, the Chittagong Hill Tracts of Bangladesh, and other parts of Northeast India, Kokborok holds official language status in Tripura alongside Bengali. It belongs to the Bodo-Garo branch of the Tibeto-Burman language family within the larger Sino-Tibetan family, and is characterised by SOV word order, postpositions, and tonal phonology.

Despite its official status and substantial speaker population, Kokborok remains severely under-resourced in natural language processing. Prior computational work has been largely limited to morphological analysis \citep{debbarma2012morphological}, POS tagging, and rule-based named entity recognition. Machine translation for Kokborok has received minimal attention: the WMT Low-Resource Indic Language Translation shared tasks \citep{pakray2023wmt,pakray2024wmt,pakray2025wmt} have included Kokborok since 2023, providing the only published NMT baselines. The best WMT 2025 submission \citep{anvita2025} achieved BLEU scores of 6.99 (en$\rightarrow$trp) and 2.99 (trp$\rightarrow$en). These low scores reflect the extreme data scarcity and domain restriction of available training data, rather than any inherent untranslatability of the language.

In this paper, we address this gap with the following contributions:

\begin{itemize}
    \item We develop KokborokMT, a significantly improved NMT system for Kokborok, by fine-tuning NLLB-200 \citep{nllbteam2022} with a novel \texttt{trp\_Latn} language token.
    \item We construct a 36,052-sentence parallel corpus combining professional translations from SMOL \citep{caswell2025smol}, WMT Bible data, and synthetic back-translations generated from Tatoeba English sentences using Gemini Flash.
    \item We demonstrate that synthetic data augmentation via LLM back-translation yields consistent improvements across all evaluation metrics and test conditions.
    \item We provide comprehensive ablation studies comparing zero-shot NLLB, fine-tuning without synthetic data (System 1), and fine-tuning with synthetic data (System 2).
    \item We investigate LaBSE-based quality filtering for synthetic data and report that filtering scores are unreliable for Kokborok due to the language's absence from LaBSE's training data, an important negative finding for the community.
    \item We conduct human evaluation with three annotators including a linguistic expert and a domain specialist, yielding mean scores of 3.74/5 (adequacy) and 3.70/5 (fluency).
    \item We release the model and evaluation scripts publicly to facilitate further research on Kokborok NLP.
\end{itemize}

\section{Background and Related Work}

\subsection{Kokborok: Language and Script}

Kokborok has a native script called Koloma, used during the reign of the Tripura kings, which is currently undergoing a revival effort. However, contemporary digital use and NLP research overwhelmingly employs the Roman script, which we adopt throughout this work. The language has nine dialects named after tribal communities (Debbarma, Reang, Jamatia, Noatia, etc.). Kokborok is an SOV language, uses suffixes as tense markers (\textit{-o} for present, \textit{-kha} for past, \textit{-nai} for future), and employs two phonemic tones (level and high). Adjectives follow the nouns they modify, and plural markers appear sentence-finally on nouns.

\subsection{Prior NLP Work on Kokborok}

Computational work on Kokborok has been sparse. \citet{debbarma2012morphological} developed a morphological analyser achieving approximately 80\% accuracy. POS taggers using CRF and SVM approaches have been reported at around 84\% accuracy. A rule-based NER system achieved 83\% F-score \citep{debbarma2019ner}. Vowel recognition using LPCC features has also been explored. For machine translation, the WMT Low-Resource Indic Language Translation shared task has included Kokborok since 2023 \citep{pakray2023wmt,pakray2024wmt,pakray2025wmt}, providing the only published NMT baselines. The best WMT 2025 submission \citep{anvita2025} achieved BLEU of 6.99 (en$\rightarrow$trp) and 2.99 (trp$\rightarrow$en). We also note that OPUS and HuggingFace contain no publicly available Kokborok parallel data beyond the WMT shared task releases, confirming the extreme scarcity of resources for this language.

\subsection{Low-Resource MT and Back-Translation}

Back-translation \citep{sennrich2016improving} is a well-established technique for augmenting low-resource MT training data by translating monolingual target-side text into the source language. Recent work has demonstrated that LLMs can serve as effective generators of synthetic parallel data for low-resource settings. The NLLB-200 model \citep{nllbteam2022} has become a standard backbone for low-resource MT fine-tuning, covering 200 languages with strong multilingual representations that transfer well to unseen languages through continued training. Adding new language tokens to multilingual models and fine-tuning on target language data is an established approach for extending coverage beyond the original training set. Fine-tuning NLLB on domain-specific or language-specific data consistently improves over zero-shot performance for languages at the margins of its coverage.

\section{Data}

\subsection{Parallel Corpus Construction}

Our training corpus combines three sources, totalling 36,052 sentence pairs after preprocessing and deduplication.

\paragraph{SMOL (9,284 sentences).} The SMOL dataset \citep{caswell2025smol} provides professionally human-translated parallel data for 123 low-resource languages. For Kokborok, SMOL comprises two sub-datasets: SMOLDOC (6,016 sentences), consisting of LLM-generated English documents on diverse topics professionally translated into Kokborok; and GATITOS (4,211 sentences), a token-level resource. An additional 57 SMOLSENT sentences were found to have reversed source and target columns during preprocessing and were corrected prior to use. SMOL represents the highest-quality component of our training data, covering diverse domains including health, education, culture, technology, and everyday conversation.

\paragraph{WMT Bible Corpus (1,769 sentences).} The WMT Low-Resource Indic Language Translation shared task \citep{pakray2025wmt} provides 2,269 Kokborok-English parallel sentence pairs derived from Bible translations. We reserve 500 sentences as a held-out test set and use the remaining 1,769 for training. While this corpus is domain-restricted, it provides additional training signal and enables direct comparison with prior WMT baselines on the same test distribution.

\paragraph{Synthetic Back-Translation (24,999 sentences).} We generate synthetic en$\rightarrow$trp parallel data using English source sentences from the Tatoeba project (\texttt{agentlans/tatoeba-english-translations} on HuggingFace) \citep{tiedemann2020tatoeba}, a well-known multilingual sentence collection used widely in MT research. We apply a length filter retaining sentences between 5 and 20 words, followed by deduplication using exact string matching, yielding 25,000 unique English sentences. Kokborok translations were generated in batches using the Google Gemini Flash API (\texttt{gemini-2.5-flash-preview} model) with the following system instruction: \textit{``You are a professional English to Kokborok translator. Translate each line accurately. Maintain the line order. Output ONLY the translations.''} The total API cost was approximately INR 600 (USD \$7). Tatoeba sentences cover everyday conversational and factual domains, complementing the more formal register of SMOLDOC.

\subsection{Quality Filtering Investigation}

Following standard practice in synthetic MT pipelines, we investigated LaBSE-based quality filtering \citep{feng2022labse} to identify and remove low-quality synthetic pairs. We computed cosine similarity between LaBSE embeddings of English source and Kokborok target sentences across all 24,999 pairs. The resulting score distribution (mean: 0.287, std: 0.216) was substantially lower than typical cross-lingual similarity scores for supported languages. Manual inspection of pairs across all score ranges confirmed that translation quality was consistently acceptable even at very low similarity scores (e.g., 0.04--0.15). We attribute the low scores to Kokborok's absence from LaBSE's training languages, rendering the embeddings unreliable for cross-lingual alignment with Kokborok. We therefore retain all 24,999 synthetic pairs without filtering and report this as a cautionary finding: LaBSE-based quality filtering is not applicable to languages absent from the model's training data.

\subsection{Data Splits and Deduplication}
\label{sec:splits}

We construct evaluation sets from the two highest-quality sources:

\begin{itemize}
    \item \textbf{SMOL Test Set (500 sentences):} Randomly sampled from SMOLDOC sentences only, ensuring domain diversity and professional translation quality.
    \item \textbf{WMT Test Set (499 sentences):} Randomly sampled from the WMT Bible corpus, enabling direct comparison with WMT shared task results.
    \item \textbf{Development Set (500 sentences):} Randomly sampled from remaining SMOL sentences after test extraction.
\end{itemize}

Zero overlap between training data (including synthetic pairs) and all test sets is verified by exact English-side string matching prior to any model training. The final training set contains 36,052 sentence pairs across all three sources.

\subsection{Data Statistics}

Table~\ref{tab:data} summarises the corpus composition. Figure~\ref{fig:overview} (right) shows the relative contribution of each source.

\begin{table}[h]
\centering
\small
\begin{tabular}{lrr}
\toprule
\textbf{Source} & \textbf{Sentences} & \textbf{Type} \\
\midrule
SMOL (train) & 9,284 & Professional \\
WMT Bible (train) & 1,769 & Professional \\
Gemini BT (Tatoeba) & 24,999 & Synthetic \\
\midrule
\textbf{Total Train} & \textbf{36,052} & \\
Dev (SMOL) & 500 & Professional \\
Test (SMOL) & 500 & Professional \\
Test (WMT) & 499 & Professional \\
\bottomrule
\end{tabular}
\caption{Corpus statistics for KokborokMT.}
\label{tab:data}
\end{table}

\section{Methodology}

\subsection{Base Model and Language Token}

We fine-tune \texttt{facebook/nllb-200-distilled-600M} \citep{nllbteam2022}, a 600M parameter sequence-to-sequence transformer covering 200 languages. Kokborok is not among NLLB's supported languages. We add a new special token \texttt{trp\_Latn} to the tokenizer vocabulary (assigned ID 256204) and resize the model's embedding matrix accordingly. This allows the model to condition generation on Kokborok as a distinct target language while leveraging representations from typologically related Tibeto-Burman languages already present in NLLB such as Burmese and Tibetan.

\subsection{Training Setup}

We train both translation directions simultaneously by concatenating the original and direction-flipped datasets, yielding 72,096 training pairs. This joint training approach encourages shared representations for both directions.

\paragraph{Hyperparameters.} We train for 10 epochs using the AdamW optimiser with learning rate 2e-5, linear warmup over 500 steps, weight decay 0.01, and batch size 32. Mixed precision training (fp16) is used throughout. Maximum sequence length is 128 tokens. Training is conducted on a single A40 GPU, completing in approximately 3.5 hours for System 2 and 1.1 hours for System 1.

\paragraph{Model Selection.} Checkpoints are saved at each epoch and the best model is selected based on validation loss on the SMOL development set. For both systems, validation loss continued to decrease through epoch 10, with final checkpoints selected (System 2 val loss: 0.2422; System 1 val loss: 0.2278).

\subsection{Experimental Conditions}

We evaluate three systems:

\begin{itemize}
    \item \textbf{Zero-Shot NLLB:} The base NLLB-200-distilled-600M model with \texttt{trp\_Latn} token added but no fine-tuning.
    \item \textbf{System 1 (No BT):} Fine-tuned on SMOL + WMT only (11,053 pairs; 23,098 with both directions). No synthetic data.
    \item \textbf{System 2 (Full):} Fine-tuned on SMOL + WMT + Gemini synthetic data (36,052 pairs; 72,096 with both directions). Primary system.
\end{itemize}

\section{Evaluation}

\subsection{Automatic Metrics}

We evaluate using a comprehensive suite of automatic metrics matching the WMT shared task evaluation protocol \citep{pakray2025wmt}:

\begin{itemize}
    \item \textbf{BLEU} \citep{papineni2002bleu}: Computed using sacreBLEU \citep{post2018call} with default tokenisation.
    \item \textbf{chrF} \citep{popovic2015chrf}: Character n-gram F-score via sacreBLEU.
    \item \textbf{ROUGE-L} \citep{lin2004rouge}: Longest common subsequence F-measure.
    \item \textbf{METEOR} \citep{banerjee2005meteor}: Alignment-based metric using WordNet.
    \item \textbf{TER} \citep{snover2006ter}: Translation edit rate (lower is better).
    \item \textbf{Cosine Similarity:} Semantic similarity via LaBSE embeddings \citep{feng2022labse}.
    \item \textbf{COMET} \citep{rei2020comet}: Neural evaluation using \texttt{Unbabel/wmt22-comet-da}.
\end{itemize}

All systems use beam search with beam size 4. We evaluate both directions on both test sets (four conditions per system).

\subsection{Human Evaluation}

\paragraph{Data.} Human evaluation was conducted on 50 en$\rightarrow$trp translations generated by System 2. Source sentences were sampled from the Tatoeba dataset \citep{tiedemann2020tatoeba} (\texttt{agentlans/tatoeba-english-translations}, HuggingFace), separately from all training and automatic evaluation data. From 6,765,220 total rows, we first deduplicated on the English column (1,417,346 unique sentences), then filtered for readability $\geq$ 2.5 and quality $\geq$ 0.0, yielding 222,317 candidate sentences. We randomly sampled 50 sentences (random\_state=99) covering everyday conversational and factual domains.

\paragraph{Annotators.} Three annotators independently rated all 50 translations: (1) a linguistic expert with expertise in Kokborok language structure; (2) a native Kokborok speaker; and (3) a native researcher specialising in Kokborok linguistics. Annotators were not informed of each other's ratings.

\paragraph{Criteria.} Each translation was rated on two 1--5 scales: \textbf{Adequacy} (does the translation preserve the meaning of the source?) and \textbf{Fluency} (is the translation natural and grammatically correct?), following standard MT human evaluation practice.

\paragraph{Results.} Table~\ref{tab:human} presents individual and aggregate scores. Agreement between the two trained evaluators (linguistic expert and native researcher) was substantial ($\kappa = 0.67$ for both adequacy and fluency), while the untrained native speaker showed lower agreement ($\kappa = 0.13$), consistent with known scale interpretation variability among non-expert annotators in MT human evaluation \citep{graham2013continuous}. Mean scores across all annotators of 3.74/5 (adequacy) and 3.70/5 (fluency) indicate that KokborokMT successfully preserves meaning in most cases and produces largely natural output.

\begin{table}[h]
\centering
\small
\begin{tabular}{lcc}
\toprule
\textbf{Annotator} & \textbf{Adequacy} & \textbf{Fluency} \\
\midrule
Linguistic expert & 3.76 & 3.76 \\
Native speaker & 3.64 & 3.54 \\
Native researcher (Kokborok) & 3.84 & 3.80 \\
\midrule
\textbf{Mean} & \textbf{3.74} & \textbf{3.70} \\
\midrule
$\kappa$ (expert vs researcher) & 0.672 & 0.677 \\
$\kappa$ (expert vs native) & 0.134 & 0.109 \\
$\kappa$ (native vs researcher) & 0.004 & 0.019 \\
\bottomrule
\end{tabular}
\caption{Human evaluation results (1--5 scale, $n=50$). Cohen's $\kappa$ reported for each annotator pair.}
\label{tab:human}
\end{table}

\subsection{Reproducibility}

Full sacreBLEU signatures for System 2 are provided in Appendix~\ref{sec:sacrebleu} for exact reproducibility.

\subsection{Automatic Evaluation Results}

\begin{figure*}[t]
  \centering
  \includegraphics[width=\textwidth]{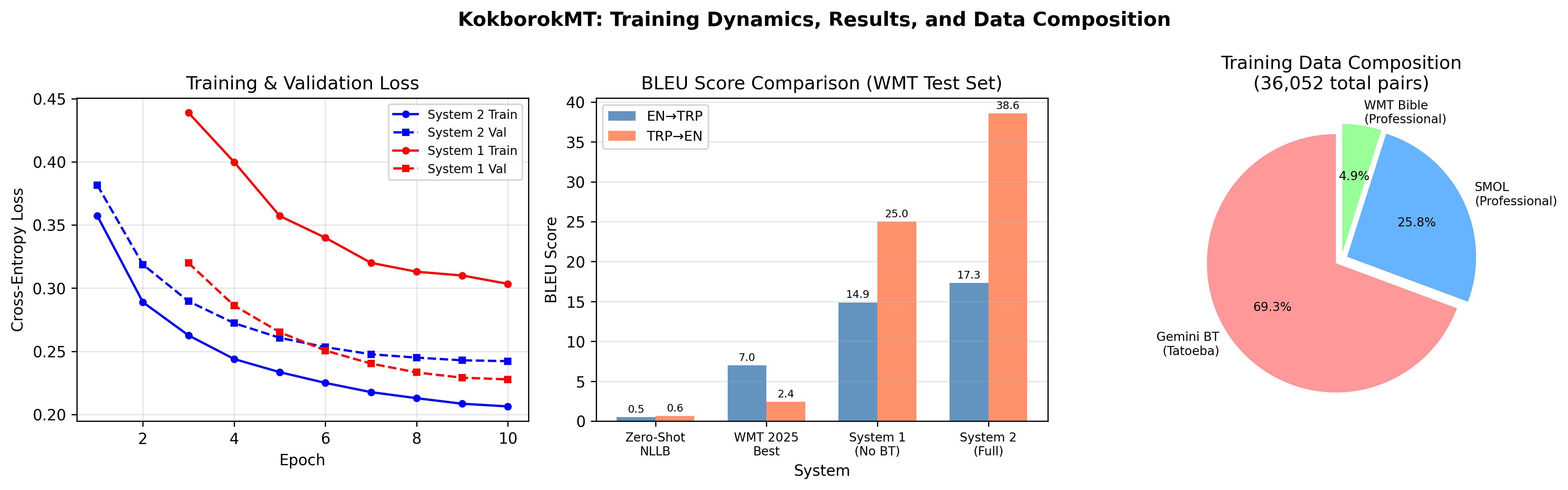}
  \caption{Left: Training and validation loss curves for System 1 (no synthetic data, red) and System 2 (full pipeline, blue). System 1 shows instability at epoch 1 due to smaller dataset size but converges cleanly from epoch 3. Centre: BLEU score comparison across all systems on the WMT test set for both translation directions. Right: Training data composition showing contribution of each source to the 36,052 total sentence pairs.}
  \label{fig:overview}
\end{figure*}

Table~\ref{tab:results} presents full automatic evaluation results. Figure~\ref{fig:overview} (centre) visualises BLEU scores across systems.

\begin{table*}[t]
\centering
\small
\begin{tabular}{llccccccc}
\toprule
\textbf{System} & \textbf{Direction} & \textbf{BLEU} & \textbf{chrF} & \textbf{ROUGE-L} & \textbf{METEOR} & \textbf{TER$\downarrow$} & \textbf{Cos Sim} & \textbf{COMET} \\
\midrule
\multicolumn{9}{l}{\textit{SMOL Test Set (general domain)}} \\
Zero-Shot NLLB & en$\rightarrow$trp & 0.50 & 11.89 & 0.0261 & 0.0132 & 139.51 & 0.1939 & 0.2697 \\
Zero-Shot NLLB & trp$\rightarrow$en & 0.63 & 17.07 & 0.0675 & 0.0526 & 130.30 & 0.1872 & 0.2880 \\
System 1 (No BT) & en$\rightarrow$trp & 13.35 & 46.22 & 0.3873 & 0.3112 & 75.95 & 0.7707 & 0.6938 \\
System 1 (No BT) & trp$\rightarrow$en & 32.91 & 49.41 & 0.5498 & 0.5091 & 55.07 & 0.7466 & 0.6604 \\
System 2 (Full) & en$\rightarrow$trp & \textbf{15.25} & \textbf{47.67} & \textbf{0.3896} & \textbf{0.3138} & \textbf{74.26} & 0.7596 & \textbf{0.6958} \\
System 2 (Full) & trp$\rightarrow$en & \textbf{38.56} & \textbf{53.92} & \textbf{0.5919} & \textbf{0.5602} & \textbf{50.15} & \textbf{0.7911} & \textbf{0.6926} \\
\midrule
\multicolumn{9}{l}{\textit{WMT Test Set (Bible domain)}} \\
Zero-Shot NLLB & en$\rightarrow$trp & 0.09 & 12.76 & 0.0136 & 0.0056 & 121.59 & 0.3361 & 0.2545 \\
Zero-Shot NLLB & trp$\rightarrow$en & 0.32 & 16.89 & 0.0560 & 0.0390 & 123.30 & 0.2701 & 0.2888 \\
System 1 (No BT) & en$\rightarrow$trp & 14.87 & 42.34 & 0.3908 & 0.3136 & 86.11 & 0.7268 & 0.6718 \\
System 1 (No BT) & trp$\rightarrow$en & 24.99 & 45.14 & 0.5078 & 0.4306 & 70.23 & 0.7889 & 0.6413 \\
System 2 (Full) & en$\rightarrow$trp & \textbf{17.30} & \textbf{47.11} & \textbf{0.4332} & \textbf{0.3483} & \textbf{76.81} & \textbf{0.7479} & \textbf{0.7064} \\
System 2 (Full) & trp$\rightarrow$en & \textbf{28.03} & \textbf{48.18} & \textbf{0.5449} & \textbf{0.4713} & \textbf{66.31} & \textbf{0.8171} & \textbf{0.6640} \\
\midrule
\multicolumn{9}{l}{\textit{WMT 2025 Shared Task Best Systems \citep{pakray2025wmt} (Bible test set only; other metrics not reported)}} \\
WMT Best & en$\rightarrow$trp & 6.99 & 38.08 & 0.367 & 0.300 & 76.26 & -- & -- \\
WMT Best & trp$\rightarrow$en & 2.99 & 25.52 & 0.218 & 0.163 & 117.73 & 0.487 & -- \\
\bottomrule
\end{tabular}
\caption{Automatic evaluation results. System 2 (Full) is our primary system. TER is lower-is-better; all other metrics are higher-is-better. WMT 2025 best system results are from \citet{pakray2025wmt} and were trained exclusively on Bible-domain data; our systems additionally use SMOL professional translations, making direct comparison indicative rather than strictly controlled.}
\label{tab:results}
\end{table*}

\section{Analysis}

\subsection{Impact of Fine-Tuning}

Zero-shot NLLB produces near-zero BLEU scores (0.09--0.63) for both directions, confirming that Kokborok lies outside NLLB's effective coverage despite the addition of the \texttt{trp\_Latn} token. Fine-tuning with even a modest gold corpus (System 1, $\sim$11k pairs) produces dramatic improvements, reaching BLEU 32.91 (trp$\rightarrow$en, SMOL) and 14.87 (en$\rightarrow$trp, WMT). This represents a roughly 30$\times$ improvement over zero-shot, demonstrating the critical importance of even modest amounts of high-quality parallel data for languages absent from multilingual model coverage.

\subsection{Impact of Synthetic Back-Translation}

System 2 consistently outperforms System 1 across all four evaluation conditions and all metrics. The gains are most pronounced for trp$\rightarrow$en on the SMOL test set (+5.65 BLEU), where the model benefits most from the additional Tatoeba-derived training signal. For en$\rightarrow$trp, gains are more modest (+1.90 BLEU on SMOL, +2.43 on WMT), reflecting the asymmetric nature of our synthetic data: Gemini translated English $\rightarrow$ Kokborok, so the synthetic data primarily strengthens the en$\rightarrow$trp direction at training time, yet the bidirectional training setup propagates improvements to both directions. The consistent improvements validate Gemini Flash as a practical source of augmentation data for extremely low-resource languages at minimal cost.

\subsection{LaBSE Quality Filtering}

We investigated LaBSE-based quality filtering on the 24,999 synthetic pairs. Mean cross-lingual cosine similarity was 0.287 (std: 0.216), far below typical values for supported language pairs. At thresholds of 0.3, 0.4, 0.5, and 0.6, only 44.6\%, 32.8\%, 20.0\%, and 10.0\% of pairs would be retained respectively. Manual inspection confirmed translation quality was acceptable across all score ranges, including pairs with similarity scores below 0.1. We conclude that LaBSE similarity scores are not a reliable quality signal for Kokborok due to its absence from LaBSE's training data. This is an important negative finding: quality filtering methods that rely on multilingual embeddings should be validated on a per-language basis before application to truly unseen languages.

\subsection{Human Evaluation Analysis}

The human evaluation scores of 3.74/5 (adequacy) and 3.70/5 (fluency) are consistent with the automatic metric performance. The native researcher in Kokborok linguistics gave the highest scores (3.84/3.80), while the untrained native speaker gave the lowest (3.64/3.54), a pattern commonly observed in MT human evaluation where expert annotators tend to apply scales more consistently \citep{graham2013continuous}. The strong agreement between the two trained evaluators ($\kappa = 0.67$) provides confidence in the reliability of the human assessment. Overall, the scores indicate the system is practically useful --- preserving most meaning and producing largely natural output --- while acknowledging room for improvement, particularly in fluency.

\subsection{Domain Generalisation}

Both systems achieve competitive scores on both the general-domain SMOL test set and the Bible-domain WMT test set, demonstrating reasonable cross-domain generalisation despite training on data from both domains. The higher trp$\rightarrow$en BLEU scores on SMOL (38.56) versus WMT (28.03) reflect domain match with the SMOL training component.

\subsection{Translation Direction Asymmetry}

trp$\rightarrow$en consistently outperforms en$\rightarrow$trp across all systems and test sets. System 2 achieves BLEU 38.56 (trp$\rightarrow$en) versus 15.25 (en$\rightarrow$trp) on the SMOL test set. This asymmetry is expected: translating into English benefits from NLLB's strong English generation capabilities, while generating Kokborok requires producing a low-resource language with limited representation in the base model. Closing this gap likely requires more gold Kokborok-side training data and potentially monolingual continued pretraining.

\subsection{Comparison with Prior Work}

Our System 2 achieves BLEU 17.30 (en$\rightarrow$trp) versus the WMT 2025 best of 6.99, and BLEU 38.56 (trp$\rightarrow$en) versus 2.99. We note that this comparison is not strictly controlled: WMT systems trained exclusively on Bible-domain data, while our systems additionally use SMOL professional translations covering diverse domains. The performance gap reflects both the richer training data and the effectiveness of our fine-tuning approach. Even on the WMT Bible test set alone, our system substantially outperforms WMT baselines, suggesting that the improvements are not solely attributable to domain match.

\section{Limitations}

This work has several limitations. First, our synthetic back-translation data was generated from English source sentences only (en$\rightarrow$trp direction via Gemini), limiting Kokborok-source diversity in synthetic data. Future work could generate trp$\rightarrow$en synthetic data from monolingual Kokborok text. Second, automatic metrics have known limitations for low-resource languages where reference translations may have limited lexical overlap with model outputs; our human evaluation partially addresses this concern. Third, we evaluate only on Roman-script Kokborok; the Bengali-script variant used in some official contexts is not addressed. Fourth, we do not attempt tokenizer adaptation or monolingual continued pretraining for Kokborok, which may further improve generation quality. Fifth, our comparison with WMT 2025 systems is indicative rather than strictly controlled due to differences in training data. Sixth, human evaluation was conducted only for the en$\rightarrow$trp direction; trp$\rightarrow$en human evaluation is left for future work.

\section{Conclusion}

We have presented KokborokMT, a significantly improved neural machine translation system for Kokborok, achieving BLEU 17.30 (en$\rightarrow$trp) and 38.56 (trp$\rightarrow$en) and substantially surpassing prior published results. Human evaluation by three annotators confirms practical translation quality with mean adequacy 3.74/5 and fluency 3.70/5. Our work demonstrates that a combination of professionally translated data from SMOL, domain-specific parallel data, and LLM-generated synthetic back-translations can bootstrap effective MT for an extremely low-resource Tibeto-Burman language. We additionally report that LaBSE-based quality filtering is unreliable for languages absent from the model's training data, a cautionary finding for the community. We will release our model and evaluation pipeline publicly under CC-BY-4.0 to support further research on Kokborok and other under-resourced languages of Northeast India.

\section*{Acknowledgments}

We thank the SMOL team at Google for releasing high-quality Kokborok translations, and the WMT shared task organisers for maintaining the Low-Resource Indic Language Translation benchmark. We are grateful to our human evaluators for their careful assessments.

\section*{Ethical Considerations}
All datasets used in this work are publicly available under open licenses: SMOL (CC-BY), Tatoeba (CC-BY), and WMT Bible data (public domain). Synthetic back-translations were generated using the Google Gemini Flash API in compliance with its terms of service for research use. Human evaluators participated voluntarily and are anonymised in this paper. The model will be released under CC-BY-4.0 upon acceptance. We acknowledge that MT systems for low-resource languages may produce errors that could mislead users; we recommend human review for critical applications.

\bibliography{custom}

\appendix

\section{sacreBLEU Signatures}
\label{sec:sacrebleu}

The following sacreBLEU signatures are provided for exact reproducibility of System 2 results:

\begin{itemize}
    \item EN$\rightarrow$TRP (SMOL): \texttt{BLEU = 15.25 47.9/20.4/10.2/5.5 (BP=1.000 ratio=1.009 hyp\_len=8142 ref\_len=8068)}
    \item EN$\rightarrow$TRP (WMT): \texttt{BLEU = 17.30 46.9/22.6/12.0/7.4 (BP=0.988 ratio=0.988 hyp\_len=12733 ref\_len=12884)}
    \item TRP$\rightarrow$EN (SMOL): \texttt{BLEU = 38.56 65.5/43.1/32.2/24.7 (BP=0.997 ratio=0.997 hyp\_len=8595 ref\_len=8620)}
    \item TRP$\rightarrow$EN (WMT): \texttt{BLEU = 28.03 58.2/33.8/21.7/14.5 (BP=1.000 ratio=1.010 hyp\_len=14381 ref\_len=14243)}
\end{itemize}

COMET scores computed using \texttt{Unbabel/wmt22-comet-da} via \texttt{unbabel-comet}. LaBSE embeddings via \texttt{sentence-transformers} library.

\section{Training Loss Curves}
\label{sec:training}

\begin{table}[h]
\centering
\small
\begin{tabular}{ccc}
\toprule
\textbf{Epoch} & \textbf{Train Loss} & \textbf{Val Loss} \\
\midrule
\multicolumn{3}{l}{\textit{System 2 (Full, With BT)}} \\
1 & 0.3573 & 0.3816 \\
2 & 0.2889 & 0.3187 \\
3 & 0.2626 & 0.2896 \\
4 & 0.2439 & 0.2723 \\
5 & 0.2335 & 0.2607 \\
6 & 0.2250 & 0.2533 \\
7 & 0.2177 & 0.2477 \\
8 & 0.2129 & 0.2450 \\
9 & 0.2085 & 0.2429 \\
10 & 0.2064 & 0.2422 \\
\midrule
\multicolumn{3}{l}{\textit{System 1 (No BT)}} \\
1 & 8.4454 & 1.4459 \\
2 & 1.8205 & 0.3779 \\
3 & 0.4388 & 0.3199 \\
4 & 0.3998 & 0.2862 \\
5 & 0.3571 & 0.2651 \\
6 & 0.3399 & 0.2506 \\
7 & 0.3200 & 0.2403 \\
8 & 0.3130 & 0.2333 \\
9 & 0.3100 & 0.2291 \\
10 & 0.3034 & 0.2278 \\
\bottomrule
\end{tabular}
\caption{Training and validation loss curves for both fine-tuned systems.}
\label{tab:loss}
\end{table}

\section{Human Evaluation Data Pipeline}
\label{sec:humandata}

Human evaluation sentences were sampled as follows: source dataset \texttt{agentlans/tatoeba-english-translations} (HuggingFace) \citep{tiedemann2020tatoeba}, 6,765,220 total rows; after deduplication on the English column: 1,417,346 unique sentences; after filtering (readability $\geq$ 2.5, quality $\geq$ 0.0): 222,317 sentences; random sample of 50 sentences (\texttt{random\_state=99}). Domain: general/everyday (conversation, facts, opinions, geography, daily life). These sentences are entirely separate from all training, development, and automatic evaluation splits.

\end{document}